\documentclass[10pt]{article} 
\usepackage[preprint]{tmlr}

\usepackage{amsmath,amsfonts,bm}

\def\eqref#1{equation~\ref{#1}}

\def\1{\bm{1}}

\DeclareMathAlphabet{\mathsfit}{\encodingdefault}{\sfdefault}{m}{sl}
\SetMathAlphabet{\mathsfit}{bold}{\encodingdefault}{\sfdefault}{bx}{n}

\usepackage{hyperref}
\usepackage{url}

\usepackage{amsmath}
\usepackage{graphicx}
\usepackage{subcaption}

\title{Natural Evolution Strategies as a Black Box Estimator for
Stochastic Variational Inference}

\author{\name Ahmad Ayaz Amin \email ahmadayaz.amin@torontomu.ca \\
      \addr Department of Computer Science\\
      Toronto Metropolitan University
      }

\def\month{MM}  
\def\year{YYYY} 
\def\openreview{\url{https://openreview.net/forum?id=XXXX}} 

\begin{document}

\maketitle

\begin{abstract}
Stochastic variational inference and its derivatives in the form of variational autoencoders enjoy the ability to perform Bayesian inference on large datasets in an efficient manner. However, performing inference with a VAE requires a certain design choice (i.e. reparameterization trick) to allow unbiased and low variance gradient estimation, restricting the types of models that can be created. To overcome this challenge, an alternative estimator based on natural evolution strategies is proposed. This estimator does not make assumptions about the kind of distributions used, allowing for the creation of models that would otherwise not have been possible under the VAE framework.
\end{abstract}

\section{Introduction}

Consider the problem of computing the posterior $p(z|x)=\frac{p(x|z)p(z)}{p(x)}$. Computing this directly is intractable for all but very simple distributions. The variational inference approach approximates $p(z|x)$ using a surrogate distribution $q(z)$ by minimizing the following KL divergence:

\begin{equation}
    D_{KL}(q({z)\parallel p(z|x)) =\ln p(x) + \mathbb E_q \left[\ln \frac{q({z})}{p(x,z)}\right]}
\end{equation}

Where $\mathbb E_q \left[\ln \frac{q({z})}{p(x,z)}\right]$ is the evidence lower bound (ELBO).

The stochastic variational inference (SVI) \citep{JMLR:v14:hoffman13a} approach to minimizing the KL divergence involves updating the model with respect to the gradient of the ELBO. In \cite{https://doi.org/10.48550/arxiv.1312.6114}, SVI is extended to allow use with automatic differentiation, yielding the stochastic gradient variational Bayes (SGVB) estimator. 

The SGVB estimator scales to large datasets very well thanks to a key mechanism known as the reparameterization trick, which allows an unbiased gradient estimate to be taken by reformulating the stochastic sample $z$ with respect to the model parameters. However, this estimator has certain flaws that restrict the types of models we wish to create.

\subsection{Problems with SGVB}

The reparameterization trick \citep{https://doi.org/10.48550/arxiv.1312.6114} was conceived in order to overcome the limitations of gradient estimation in SGVB. However, this estimator is only available for certain continuous distributions (e.g. the Gaussian distribution). For all other distributions, the score function estimator (REINFORCE) is used instead. While this estimator is general, it comes at the cost of having very high variance, resulting in slow and unstable training even with control variates \citep{hewitt2020learning}.

For discrete distributions in particular, an alternative would be to use a continuous approximation instead \citep{https://doi.org/10.48550/arxiv.1611.01144, https://doi.org/10.48550/arxiv.1611.00712}. This approach is much more stable as they utilize the reparameterization trick to their advantage. Yet, continuous approximations of discrete distributions have issues with scaling as they require evaluating all possible outcomes, making them impractical for certain problems \citep{https://doi.org/10.48550/arxiv.1805.10469}.

Yet another option is forgo the SGVB framework altogether and utilize wake-sleep methods \citep{Hinton1995TheA}. The wake-sleep algorithm is the precursor to the SGVB algorithm, where learning involves alternating between maximizing the ELBO with respect to the generative model and minimizing the reversed KL divergence $D_{KL}(p(z|x)\parallel q(z))$ with respect to the inference network at each iteration. This means updating the inference network using samples from the current iteration of the generative model, which results in a biased gradient estimate and is not in general convergent \citep{bornschein2015reweighted}.

Aside from the problems related to gradient estimation, there is also the issue with the kinds of functions we can use to model the data. For certain problems, we may be required to use non-differentiable functions, or make use of hybrid models \cite{DBLP:journals/corr/abs-2107-06393}. For such problems, the above inference methods are not sufficient, thus it is necessary to have a solution that pertains to these problems.

\subsection{Natural Evolution Strategies}
Natural evolution strategies (NES) are a class of black-box optimization algorithms based on the estimation of the natural gradient \citep{https://doi.org/10.48550/arxiv.1106.4487}. NES algorithms are promising in that they do not make any assumptions about the underlying function other than that it can be evaluated, making it suitable for functions that are are otherwise difficult to optimize using standard gradient-based optimization.

The NES algorithm used in this paper \citep{https://doi.org/10.48550/arxiv.1703.03864} is defined as the following gradient estimator:

\begin{equation}
\nabla_{\theta}\mathbb{E}_{\epsilon \sim N(0, I)}F(\theta + \sigma\epsilon) = \frac{1}{\sigma}\mathbb{E}_{\epsilon \sim N(0, I)} \{F(\theta + \sigma\epsilon)\epsilon\}
\end{equation}

Where $\theta$ are the model parameters, $\sigma$ is the scale parameter for the noise variable $\epsilon$, and $F$ is the score function to be maximized. In the context of this work, $\theta$ is the parameters of the generative model $p(x,z)$ and approximate posterior $q(z)$, and $F$ is the ELBO.

Learning a function using NES amounts to first sampling multiple random perturbation vectors $\epsilon$ from a standard normal distribution and scaling them by $\sigma$. Then, the perturbation vector is added to the $\theta$ and then evaluated under $F$. The resulting score is used to weight $\epsilon$ before finally being averaged with respect to $\epsilon$ and rescaled by $\sigma$.

The above gradient estimator is very similar to the reparameterization trick in that the gradient is evaluated with respect to the model parameters rather than the probability distribution being sampled from. Unlike the reparameterization trick, the function defining the model need not be differentiable, which is a requirement for the SGVB algorithm.

NES is highly scalable and competitive to vanilla gradient descent methods, particularly in problems where there is sparse gradients \citep{https://doi.org/10.48550/arxiv.1703.03864}. Moreover, as a gradient estimator, the variance of NES is lower and more stable than that of policy gradients (i.e. REINFORCE) \citep{https://doi.org/10.48550/arxiv.1703.03864}, making it appealing for use with SVI.

Beyond the theoretical promises offered by NES, it has been demonstrated that NES outperforms gradient descent in GAN inversion, a task closely related to SVI \citep{huh2020transforming}. With this in mind, a new gradient estimator based on NES is proposed, called Natural Evolution Strategies Variational Bayes (NESVB).

\section{Experiments}

\subsection{Noisy Scale}
This first experiment verifies the feasability of NESVB through a toy example taken from the Pyro probabilistic programming language documentation \footnote{See https://pyro.ai/examples/intro\_part\_ii.html}. The generative model and parameters specified are the exact same as the tutorial, and we use the same proposal distribution (diagonal Gaussian with mean of 8.5 and standard deviation of 1.0), observation (9.5) and inference steps (2500) as well. For NESVB, we set the number of parameter samples from the search distribution to 25 samples with mirrored sampling \citep{10.5555/1885031.1885034}. The generative process $p(\gamma|\chi)p(\chi)$ is defined as such:

\begin{gather}
  \chi \sim \mathcal{N}(8.5,1.0) \\
  \gamma \sim \mathcal{N}(\chi,0.75)
\end{gather}

The approximate posterior has the following formulation:
\begin{equation}
    \chi \sim \mathcal{N}(\theta, \phi)
\end{equation}

Where $\theta$ and $\phi$ are free parameters representing the mean and standard deviation respectively. For stability purposes, rather than optimizing the standard deviation directly, we optimize the log variance and compute the standard deviation from it. We measure the performance of NESVB using SGVB and reweighted wake sleep (RWS) with 5 particles \citep{bornschein2015reweighted} as baselines. The experiment is run 5 times and averaged. The results after inference are shown below.

\begin{figure}[ht]
  \begin{subfigure}{0.5\textwidth}
      \centering
      \includegraphics[width=0.9\linewidth]{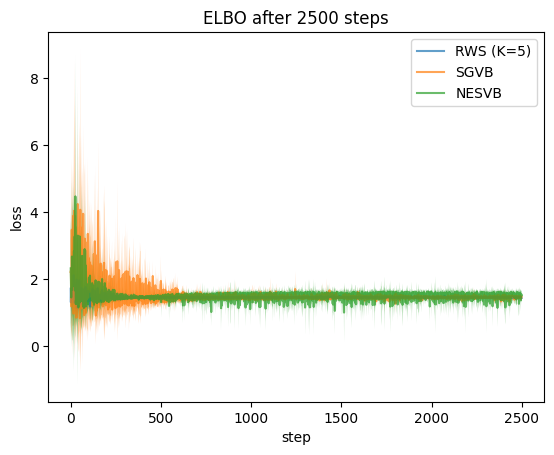}
  \end{subfigure}
  \begin{subfigure}{0.5\textwidth}
      \centering
      \includegraphics[width=0.9\linewidth]{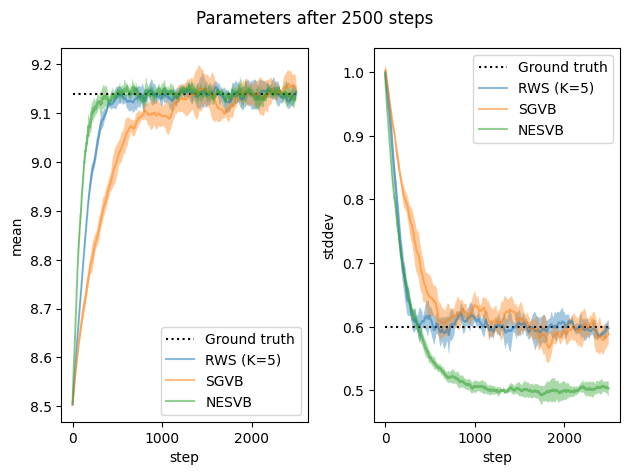}
  \end{subfigure}
  \caption{ELBO and parameters of approximate posterior}
\end{figure}

Judging from the relatively quick convergence, NESVB is very much a capable gradient estimator for SVI. It is especially intriguing given that it is tied with RWS even though it does not make use of multiple samples from the approximate posterior. With that being said, there is a slight issue in the actual model parameters, namely with the standard deviation parameter.

A similar behaviour is seen with the parameters, where NESVB and RWS have faster convergence than SGVB (albeit NESVB has slightly faster convergence). This is the case with the mean parameters. For the standard deviation parameter, NESVB slightly overshoots, although not to an extent where the approximate posterior loses all stability.

Although the exact reason for this behaviour is not entirely known, one thing which is certain is that the random sampling plays a big role in keeping the approximate posterior stable by acting as a regularizer \citep{DBLP:journals/corr/abs-1903-12436}.

To demonstrate this, the same experiment is performed but with the stochastic sampling replaced with a deterministic mapping of the mean:

\begin{figure}[ht]
  \begin{subfigure}{0.5\textwidth}
      \centering
      \includegraphics[width=0.9\linewidth]{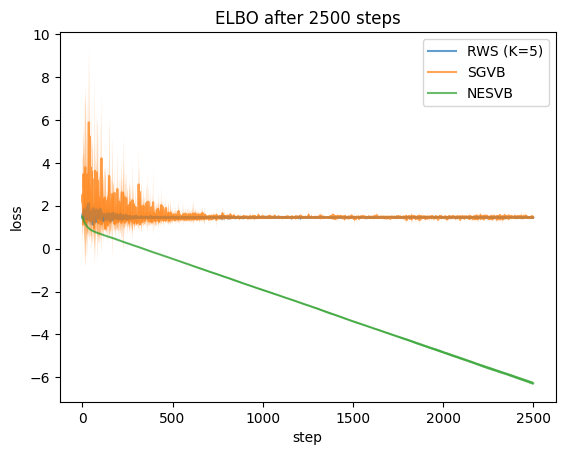}
  \end{subfigure}
  \begin{subfigure}{0.5\textwidth}
      \centering
      \includegraphics[width=0.9\linewidth]{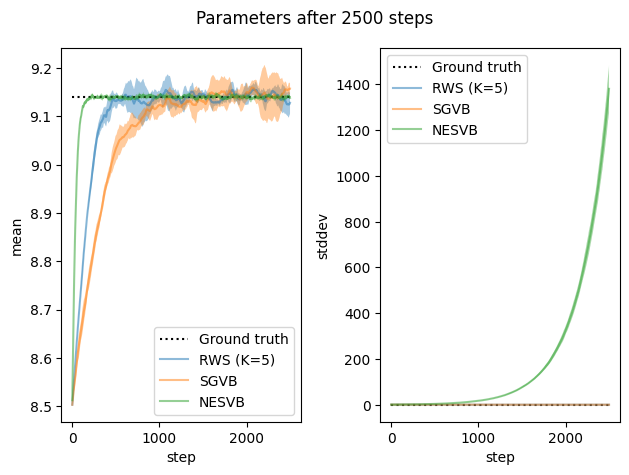}
  \end{subfigure}
  \caption{ELBO and parameters data for ablation test}
\end{figure}

In this ablation test, both SGVB and RWS converge as expected. On the other hand, NESVB shoots off into the negative ELBO zone with no sign of converging, indicating abnormality in the approximate posterior. The parameter log confirms that this is the case, where the mean converges (faster this time) but the standard deviation diverges away massively from the ground truth value.

The stochasticity of random sampling thus prevents the approximate posterior from losing meaning. Aside from this little issue, NESVB performs better than the standard gradient estimators for SVI in continuous tasks.

\subsection{GMM Clustering}
The aim of this next experiment is twofold: the first is to demonstrate that NESVB works on discrete data, and the second is to demonstrate that it can scale to larger datasets. To that end, the approximate posterior is represented by a neural network.

A synthetic 2D dataset is generated from a Gaussian Mixture Model. The GMM is composed of three Gaussians $\mathcal{N}(-1.0, 0.5)$, $\mathcal{N}(3.0, 0.25)$ and $\mathcal{N}(-5.0, 0.45)$, and is symmetric (i.e. both dimensions of the data are sampled from the same distribution).

For the baseline, the Straight-through Gumbel-Softmax gradient estimator \citep{https://doi.org/10.48550/arxiv.1611.01144} is used. Both the NESVB and Gumbel-Softmax posteriors use the same network architecture composed of a single linear layer of 2 input units (dimensions of the datapoint) and 3 hidden units (number of clusters) followed by a softmax. The output of the network represents the probabilites of each GMM component, and the component is selected by sampling from this distribution. The likelihood is computed using the GMM itself. The models are trained for 500 steps 5 times with the results averaged.

Looking at Figure 3, both NESVB and SGVB have the overall same convergence behaviour, indicating that NESVB not only works but is also competitive in the discrete domain. For the SGVB model, there is a sharp drop in loss before moving back to stable values. This behaviour may be attributed to the straight through estimator since it is not grounded in theoretical guranatees.

As for the task of clustering itself, there were some discrepencies between the two models. The NESVB model is successful in completely clustering one of the components and mostly clustering another. The SGVB model on the other hand has some trouble with separating the different clusters, indicated by the mix of classes within each cluster. From a technical standpoint, the NESVB model demonstrates stability both theoretically as well as empirically compared to the straight through estimator.

\newpage

\begin{figure}[ht]
  \begin{subfigure}{\textwidth}
      \centering
      \includegraphics[scale=0.6]{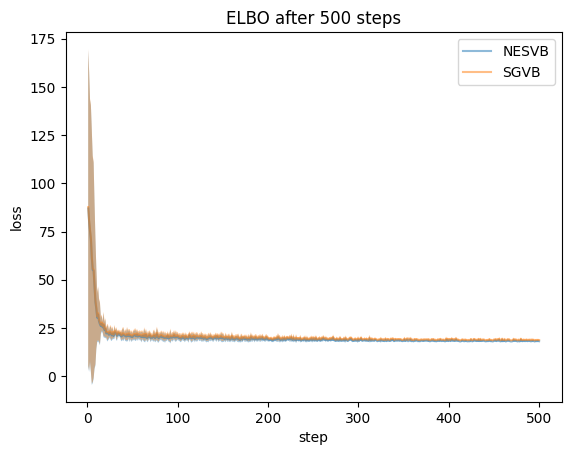}
  \end{subfigure}
  \newline
  \begin{subfigure}{\textwidth}
      \centering
      \includegraphics[scale=0.6]{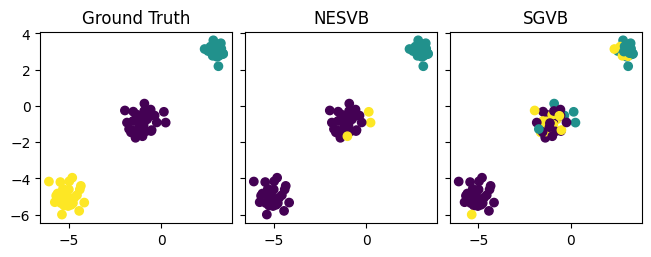}
  \end{subfigure}
  \caption{ELBO and clustering for GMM inference networks}
\end{figure}

\section{Discussion and Conclusion}
In this paper, a new gradient estimator for SVI is introduced based on natural evolution strategies. The estimator has several advantages over the current gradient descent approaches, including low variance gradient estimates, fast convergence, ability to work with any distribution, and generalization to non-differentiable models.

From a theoretical standpoint, NESVB can be seen as a Monte Carlo approximation of the true posterior since Estimation of Distribution algorithms (of which NES is a part) can be written as Monte Carlo Expectation Maximization, or exact EM in the limit of infinite samples \citep{brookes2022view}.

In terms of disadvantages, NES algorithms are slower to converge in situations where the exact gradient can be taken, such as supervised learning \citep{lenc2019nondifferentiable}. However, the results of the experiments in this study show otherwise. The unusually fast convergence of NESVB points to the fact that the solutions for SVI don't compute an exact gradient (hence gradient \textit{estimator}). This is further reinforced by the fact that both SVI and reinforcement learning (where the NES algorithm was initially used with competitive results) make use of gradient estimators to solve their problems, meaning this behaviour is justified naturally.

For now, NES is either competitive or outperforms gradient descent in SVI. In the future, NESVB may be used in conjuction with a fully differentiable model in order to combine the strengths of both NES and gradient descent approaches.

\bibliography{main}
\bibliographystyle{tmlr}

\end{document}